\documentclass[10pt,twocolumn,letterpaper]{article}

\usepackage{cvpr}
\usepackage{times}
\usepackage{epsfig}
\usepackage{graphicx}
\usepackage{amsmath}
\usepackage{amssymb}
\usepackage{mathtools}

\usepackage{multicol}

\usepackage{multirow}

\usepackage{colortbl}

\newcommand*\samethanks[1][\value{footnote}]{\footnotemark[#1]}

\usepackage{dirtytalk}
\usepackage[pagebackref=true,breaklinks=true,letterpaper=true,colorlinks,bookmarks=false]{hyperref}

\cvprfinalcopy % *** Uncomment this line for the final submission

\def\cvprPaperID{49} % *** Enter the CVPR Paper ID here

\ifcvprfinal\fi
\begin{document}

%%%%%%%%% TITLE

\title{The End-of-End-to-End: A Video Understanding Pentathlon Challenge (2020)}
\vspace{-2cm}

\author{Samuel Albanie\thanks{Equal contribution. Correspondence to albanie@robots.ox.ac.uk}, Yang Liu\samethanks, Arsha Nagrani\samethanks, Antoine Miech\samethanks, Ernesto Coto\samethanks, \\ Ivan Laptev,  Rahul Sukthankar, Bernard Ghanem, Andrew Zisserman, Valentin Gabeur,\\Chen Sun, Karteek Alahari, Cordelia Schmid, Shizhe Chen, Yida Zhao, Qin Jin, Kaixu Cui, Hui Liu,\\ Chen Wang, Yudong Jiang, Xiaoshuai Hao
}
\affiliation{Oxford}

\maketitle
%\thispagestyle{empty}

%%%%%%%%% ABSTRACT

\begin{abstract}

We present a new video understanding pentathlon challenge, an open competition held in conjunction with the IEEE Conference on Computer Vision and Pattern Recognition (CVPR) 2020. The objective of the challenge was to explore and evaluate new methods for text-to-video retrieval---the task of searching for content within a corpus of videos using natural language queries. This report summarizes the results of the first edition of the challenge together with the findings of the participants. Individual reports, dataset information, rules, and released source code can be found at the competition webpage\footnote{\url{https://www.robots.ox.ac.uk/~vgg/challenges/video-pentathlon/}}.

\end{abstract}
\section{Introduction}

Convolutional neural networks have yielded unprecedented progress on a wide range of image-centric benchmarks, driven through a combination of well-annotated datasets and end-to-end training. However, naively extending this approach from images to higher-level video understanding tasks quickly becomes prohibitive with respect to the computation and data annotation required to jointly train multi-modal high-capacity models. 

In this challenge,  we focus on an alternative “experts-driven”  approach---features are first pre-extracted from a wide range of pretrained models (the experts) and cached as an intermediate representation (specialised for semantically relevant machine perception tasks) that can then be used to train the final system. The goal of this challenge is to build a system to retrieve videos from natural language queries across a ``pentathlon" of five video retrieval benchmarks. Rather than training a retrieval system ``end-to-end", participants are provided with a diverse collection of carefully curated visual, audio and natural language pre-extracted features.

There are several benefits to the experts-driven approach:  (a) \textit{Practicality}---models for novel tasks can be composed together to exploit the available annotation in a data-efficient manner (by contrast, learning robust representations across all modalities from scratch would require vast levels of annotation to achieve comparable performance);  (b) \textit{Effectiveness}---the experts-driven approach now represents the current state-of-the-art on many video and language understanding tasks \cite{Liu19a,miech2018learning}; (c) \textit{Accessibility}---it enables researchers without access to industrial computing clusters to contribute towards questions of fundamental importance to video understanding.

This report summarizes the findings of the 2020 video understanding pentathlon challenge.  The rest of the report is structured as follows: in Sec.~\ref{sec:mechanics}, we describe the mechanics of the challenge together with the datasets that make up the pentathlon; in Sec.~\ref{sec:phase}, we describe the challenge phases and evaluation rules.  Then, in Sec.~\ref{sec:method} we offer a brief overview of the methods used by participants in the challenge and the final competition ranking, before concluding in Sec.~\ref{sec:conclusion}.   

\section{Challenge Mechanics} \label{sec:mechanics}

In this section, we describe the datasets selected to form the video pentathlon, the pre-extracted features and the baseline model provided to the participants.

% \begin{table*}[h]
% \centering 
 
%  \begin{tabular}{@{\extracolsep{4pt}}lcccc|cccc}
%  \hline \hline 
%  \multicolumn{2}{l}{} & \multicolumn{3}{c}{\textbf{Videos} \quad \quad \quad \quad \quad \quad \enspace} \multicolumn{3}{|c}{\textbf{\hspace{1cm} Queries}} & \multicolumn{1}{c}{} \\
%  \cline{2-8} %& \cline{1-3} 
%  \textbf{Dataset} & train & val & server val & server test & server val & server test & max queries per video  \\
%  \hline
 
% MSVD~\cite{chen2011collecting} & 1080 & 120 & 100 & 670& 4290 & 27763 & 81\\
% DiDeMo~\cite{anne2017localizing} &	7552 & 840 & 1065 & 1004 & 1065 & 1004 & 1 \\
% ActivityNet~\cite{krishna2017dense} & 	8007 &	1001& 1001 & 4917 & 1001 & 4917 & 1 \\
% MSRVTT~\cite{xu2016msr} &	5861 &	652 & 497	& 2990 & 9940 & 59794 & 20 \\
% YouCook2~\cite{ZhXuCoCVPR18} & 7745 & 968 & 969 & 3310 & 969 & 3310 & 1 \\
%  \hline \hline 
%  \end{tabular}
% \vspace{1em}
% \caption{Statistics of the five datasets used in the Video Pentathlon Challenge, with the partitions of each dataset. ** this table is very hard to understand. Explain what the columns mean. Add a vertical divider between semantically different blocks ** }
% \label{table:stat} 
% \end{table*}

\begin{table*}[h]
\centering 
 
 \begin{tabular}{@{\extracolsep{4pt}}lcccc|c}
 \hline \hline 
%  \multicolumn{2}{l}{} & \multicolumn{4}{c}{videos} \multicolumn{1}{|c}{Queries}\\
%  \cline{2-8} %& \cline{1-3} 
% \multirow{2}{*}{Dataset} & train & val & public\_server\_val & public\_server\_test & max queries \\&videos & videos & videos & videos & per video \\
%  \textbf{Dataset} & train \\videos & val videos & server val videos & server test videos & max queries per video  \\
%  \textbf{Dataset} & train & val & server val & server test & server val & server test & max queries per video  \\
Dataset & train vids & val vids & public\_server\_val vids & public\_server\_test vids & max queries per vid\\
 \hline
 
MSVD~\cite{chen2011collecting} & 1080 & 120 & 100 & 670& 81\\
DiDeMo~\cite{anne2017localizing} &	7552 & 840 & 1065 & 1004 & 1 \\
ActivityNet~\cite{krishna2017dense} & 	8007 &	1001& 1001 & 4917 & 1 \\
MSRVTT~\cite{xu2016msr} &	5861 &	652 & 497	& 2990 & 20 \\
YouCook2~\cite{ZhXuCoCVPR18} & 7745 & 968 & 969 & 3310 & 1 \\
 \hline \hline 
 \end{tabular}
\vspace{1em}
\caption{Statistics of the five datasets and four partitions used in the Video Pentathlon Challenge. Paired data for the train and val splits were made available for model development. Paired data for the public\_server\_val and public\_server\_test partitions was withheld and stored on an evaluation server. The former was provided to enable participants to sanity check their models, while the latter was used to produce the final ranking of the challenge (the challenge phases corresponding to these splits are described in Sec.~\ref{sec:phase}). }
\label{table:stat} 
\end{table*}

\subsection{Dataset Selection}

The \textit{video pentathlon} consisted of the five following datasets that constitute the benchmarks/challenges of the pentathlon:\\

\noindent\textbf{MSVD~\cite{chen2011collecting}:} comprises a total of 80K descriptions (in English) for 1,970 videos sourced from YouTube (with approximately 40 sentences per video). Unlike the other datasets featured in the pentathlon, the videos contained in MSVD do not possess audio streams. \\

\noindent\textbf{DiDeMo~\cite{anne2017localizing}:} consists of unedited, personal videos that are collected in an open-world setting and which include diverse content such as pets, music concerts and sports games. The dataset comprises 10,464 videos which are accompanied by approximately 3-5 pairs of descriptions and distinct moments per video. \\
    
\noindent\textbf{ActivityNet(+captions)~\cite{krishna2017dense}:} contains a total of 15K videos (sourced from the original ActivityNet dataset) accompanied by approximately 100K descriptive sentences. The videos, originally sourced from YouTube, exhibit a broad diversity of actions and content. \\
   
\noindent\textbf{MSR-VTT~\cite{xu2016msr}:} contains 10K videos sourced from YouTube which are accompanied by 200K descriptive captions (thus, there are 200K unique video-caption pairs in total). \\

\noindent\textbf{YouCook2~\cite{ZhXuCoCVPR18}:} includes 2000 long untrimmed videos from 89 cooking recipes; on average, each distinct recipe has 22 videos. The videos are sourced from YouTube and contains content filmed from a third-person viewpoint with unfixed cameras.\\

The statistics of the five datasets are provided in Table \ref{table:stat}, together with information about the train/test partitions.

\subsection{Pre-extracted Experts}
A diverse collection of carefully curated visual, audio and natural language pre-extracted features were provided to the participants including 8 features pre-extracted from visual perception models, 2 features from audio models and 2 features from natural language models. To produce features of a manageable size, the raw model outputs were temporally aggregated in three ways: (1) temporal average pooling (across frames); (2) temporal max pooling (across frames) and (3) ``fixed\_seg", where the features were partitioned into a fixed number of uniformly spaced ``chunks" (8 in total) and then average pooled within the chunk (the goal of this aggregation strategy was to preserve coarse-grained temporal information). 

Since the test set of each of the datasets was already public, the features were obfuscated prior to release. Further details on the features are provided below (for each set of features, we provide the name used to describe the features on the challenge website in brackets). \\

\noindent\textbf{Perception Models}\\

% ** give some overview here: that there are object features, scene features etc. Why there is redundancy (e.g.\ several types of object features) ** 
We provided pre-extracted visual perception features for object, scene and action recognition, as well as for face-verification and optical character recognition (OCR). For certain categories, we provide multiple models to enable retrieval systems to benefit from with different architectures or pretraining data.  \\

% \begin{itemize}
\noindent\textbf{1. Object Features (imagenet.resnext101.0):} are extracted using a ResNeXt-101 model~\cite{xie2017aggregated} that has been pretrained on Instagram hashtags~\cite{mahajan2018exploring} and fine-tuned on ImageNet for the task of image classification. Features are extracted from frames extracted at 25 fps, where each frame is resized to 224 $\times$ 224 pixels. The dimension of the embeddings is 2048 and the dimension of logits is 1000. \\

\noindent\textbf{2. Object Features (imagenet.senet154.0):} are extracted using a SENet-154 model~\cite{hu2019squeeze} that has been trained on ImageNet for the task of image classification. Features are extracted from frames extracted at 25 fps, where each frame is resized to 224 $\times$ 224 pixels. The dimension of the embeddings is 2048 and the dimension of logits is 1000. \\
    
\noindent\textbf{3. Scene Features (scene.densenet161.0):} are extracted from 224 $\times$ 224 pixel centre crops with a DenseNet-161~\cite{huang2017densely} model pretrained on Places365~\cite{zhou2017places}. The dimension of the embeddings is 2208 and the dimension of logits is 365. \\
   
\noindent\textbf{4. Action Features (i3d.i3d.0):} are extracted with an I3D inception model pretrained on Kinetics-400 that computes features following the procedure described by~\cite{carreira2017quo}. Frames are extracted at 25fps and processed in batches of 64 with a stride of 25 frames. Each frame is first resized to a height of 256 pixels (preserving aspect ratio), before a 224 $\times$ 224 centre crop is passed to the model. The dimension of the embeddings is 1024 and the dimension of logits is 400. \\

\noindent\textbf{5. Instructional Video Features (s3dg.s3dg.0):} are extracted with an S3D~\cite{xie2018rethinking} model that computes features following the learning procedure described by~\cite{miech2020end} trained on the HowTo100M dataset~\cite{miech2019howto100m}. Frames are extracted at 10fps and processed in clips of 32 frames with a stride of 16 frames. Each frame is first resized to a height of 256 pixels (preserving aspect ratio), before a 224 $\times$ 224 centre crop is passed to the model. The dimension of the embeddings is 1024 and the dimension of logits is 512. \\
    
\noindent\textbf{6. Instagram Features (r2p1d.r2p1d-ig65m.0):} are extracted with with a 34-layer R(2+1)D model~\cite{tran2018closer} trained on IG-65m~\cite{ghadiyaram2019large}  which processes clips of 8 consecutive 112 $\times$ 112 pixel frames, extracted at 30 fps (we use the implementation provided by~\cite{Daniel}). The dimension of the embeddings is 512 and the dimension of logits is 359. \\
    
\noindent\textbf{7. Instagram Video Features (r2p1d.r2p1d-ig65m-kinetics.0):} are extracted with a 34-layer R(2+1)D model~\cite{tran2018closer} trained on IG-65m~\cite{ghadiyaram2019large} and then fine-tuned on Kinetics-400 \cite{carreira2017quo} which processes clips of 8 consecutive 112 $\times$ 112 pixel frames, extracted at 30 fps (as above, we use the implementation provided by~\cite{Daniel}). The dimension of the embeddings is 512 and the dimension of logits is 400. \\
    
\noindent\textbf{8. Face features (face):} are extracted in two stages: (1) Each frame (also extracted at 25 fps) is resized to 300 $\times$ 300 pixels and passed through an SSD face detector~\cite{liu2016ssd,opencv_library} to extract bounding boxes; (2) The image region of each box is resized such that the minimum dimension is 224 pixels and a centre crop is passed through a ResNet50~\cite{he2016identity} that has been trained for the task of face classification on the VGGFace2 dataset~\cite{cao2018vggface2}, producing an embedding for each detected face. The dimension of the embeddings is 512. \\
    
\noindent\textbf{9. Optical Character Recognition Features (OCR):} are extracted in two stages: (1) Each frame is resized to 800 $\times$ 400 pixels) and passed through Pixel Link~\cite{deng2018pixellink} text detection model to extract bounding boxes for texts; (2) The image region of each box is resized to 32 $\times$ 256 and then pass these through a model~\cite{liu2018synthetically} that has been trained for scene text recognition on the Synth90K dataset~\cite{jaderberg2014synthetic}, producing a character sequence for each detect box. They are then encoded via a pretrained word2vec embedding model~\cite{mikolov2013efficient}. The dimension of the embeddings is 300 (word2vec). \\
% \end{itemize}

\noindent\textbf{Audio Models}\\
% \begin{itemize}

\noindent\textbf{1. Sound Features (audio):} are obtained with a VGGish model, trained for audio classification on the YouTube-8m dataset~\cite{hershey2017}. To produce the input for this model, the audio stream of each video is re-sampled to a 16kHz mono signal, converted to an STFT with a window size of 25ms and a hop of 10ms with a Hann window, then mapped to a 64 bin log mel-spectrogram. Finally, the features are parsed into non-overlapping 0.96s collections of frames (each collection comprises 96 frames, each of 10ms duration), which is mapped to a 128-dimensional feature vector. The dimension of the embeddings is 128. \\
    
\noindent\textbf{2. Speech Features (speech):}  The audio stream of each video is re-sampled to a 16kHz mono signal. We then obtained transcripts of the spoken speech for MSRVTT, MSVD and ActivityNet using the Google Cloud Speech to Text API from the resampled signal. The language for the API is specified as English. The dimension of the embeddings is 300 (word2vec). \\
% \end{itemize}

\noindent\textbf{Natural Language Models:} \\
% \begin{itemize}

\noindent\textbf{1. Word2Vec Features (text-w2v):} Each word of the video description is encoded using the Google News trained word2vec word embeddings ~\cite{mikolov2013efficient}. The dimension of the embeddings is 300. \\
    
\noindent\textbf{2. OpenAI Features (text-openai):} Each word of the video description is encoded with a pretrained OpenAI-GPT model~\cite{radford2018improving} to extract context-specific word embeddings (i.e., not only learned based on the current word but also the sequential context). The dimension of the embeddings is 768. \\
% \end{itemize}

\subsection{Baseline Model}
In order to provide a starting point for entrants to the challenge, we provided solid baseline code for each dataset. The baseline model provided consisted of a simple joint text-video embedding which operated on pre-computed ImageNet and I3D features, supporting the method variants described in \cite{Liu19a} and \cite{miech2018learning}. Code for the baseline model can be found at the challenge page\footnote{ \url{https://www.robots.ox.ac.uk/~vgg/challenges/video-pentathlon/challenge.html}}.

\section{Challenge Phases and Evaluation Rules} \label{sec:phase}

\begin{figure*}[ht]
\begin{center}
   \includegraphics[width=0.7\linewidth]{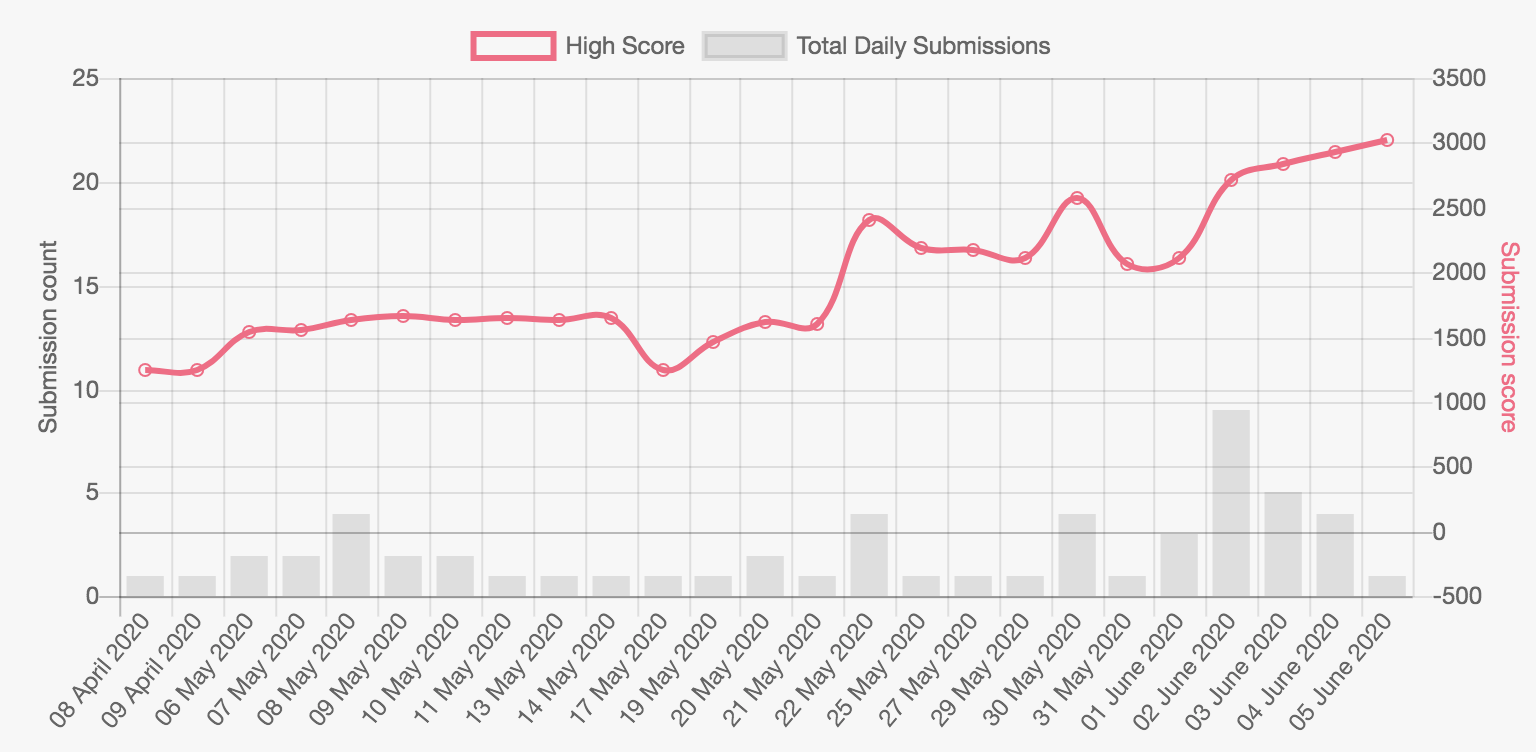}
\end{center}
   \caption{The evolution of the top leaderboard val score through time.}
   \label{fig:evolution}
\end{figure*}

\begin{table*}[h]
\centering 
 \begin{tabular}{l|c|c|c|c|c|c}
 \hline \hline 
User &	Total Score &	MSVD&	DiDeMo &	ActivityNet &	MSRVTT&	YouCook2\\
\hline
MMT&	\textbf{2511.43 (1)}	&70.24 (2)	& \textbf{46.30 (1)}&	51.57 (2)&	\textbf{70.15 (1)} &	\textbf{27.12 (1)}\\
cszhe&	2448.56 (2)&	\textbf{75.33 (1)} &	45.88 (2)&	\textbf{51.76 (1)} &	66.32 (2)	&13.76 (3)\\
acdart&	1994.89 (3)&	58.12 (4)&	33.34 (4)&	40.79 (3)&	50.99 (3)	&24.39 (2)\\
LEgGOdt	&1895.01 (4)&	59.58 (3)&	33.89 (3)&	38.29 (4)&	49.77 (4)	&9.60 (6)\\
haoxiaoshuai&	1496.98 (5)	&41.95 (6)&	31.16 (5)&	34.28 (5)&	24.55 (6)&	10.00 (4)\\
zzu	&1459.72 (6)&	42.40 (5)&	25.47 (7)&	23.30 (7)&	35.58 (5)&	9.88 (5)\\
vgg (baseline) &1250.00 (7)&	28.95 (7)&	26.06 (6)&	29.06 (6)&	14.91 (7)&	7.54 (7)\\
bland&	1249.46 (8)&	28.88 (8)&	26.06 (6)&	29.06 (6)&	14.90 (8)&	7.54 (7)\\
\hline \hline 
 \end{tabular}

\caption{Video Understanding Pentathlon Challenge 2020 final results. The number in parentheses indicates ranking and \textbf{bold text} highlights the top ranked result under each metric.}
\label{table:final_result} 
\end{table*}

Submissions were made through the CodaLab website\footnote{\url{https://competitions.codalab.org/competitions/24292}}. The challenge had two phases, corresponding to the two partitions of the data which were used for the evaluation. The two phases were:\\ 

\noindent\textbf{1. Development/Val Phase:} The `public\_server\_val' partition was open continuously throughout the challenge (from $9^{th}$ April 2020) and provided an opportunity for participants to assess progress and sanity check their submissions. This computed results on the public validation partition of each dataset.\\
    
\noindent\textbf{2. Challenge Phase:} The `public\_server\_test' was used to produce the final ranking of submissions. The challenge phase took place between $9^{th}$ May 2020 and $4^{th}$ June 2020. This computed results on the public test partition of each dataset.\\

Only one submission per day per team was allowed. In total, each team could make 30 submissions to the validation set and 3 submissions to the test set. For this challenge, participants could process the text as they wished, but training on visual features from external datasets was not permitted.

Entries into the challenge were scored under a decathlon style scoring system (inspired by its usage in the visual decathlon \cite{rebuffi2017learning}). For each of the five datasets $i \in {1,...,5}$, we first compute a measure of the quality of retrieval in each individual dataset. This ``quality measure" $g_i$ comprises the geometric mean of recall $@K$ for $K \in{1,5,10}$, computed as follows:

\begin{equation}
g_i =\Big(\prod_{k\in \{1,5,10\}} r_{i,k}\Big)^\frac{1}{3},
\end{equation}
where $r_{i,k}$ represents the recall @k on the $i^{th}$ dataset, i.e., the rate at which the correct video is retrieved amongst the top $k$ ranked results. The overall pentathlon score used for the final ranking of the submissions is then computed as follows:
\begin{equation}
S= \sum_{i=1}^{5} \alpha_i\max\{0, g_i - g_i ^{\text{offset}}\}^\gamma,
\end{equation}
where $\gamma$ is an exponential scaling factor that rewards gains in performance more heavily as they grow greater, the value is set to 2;  $g_i ^{\text{offset}}$ is a value that ensures that the baseline models achieve a score of 250 points on each dataset. The baselines, therefore, act to calibrate the difficulty of each dataset; $\alpha_i$ is assigned the value $1000(1-g_i^{\text{offset}})^{-\gamma}$, which ensures that a perfect score $g_i$ achieves a results of 1000.

\begin{figure*}[ht]
\begin{center}
   \includegraphics[width=0.8\linewidth]{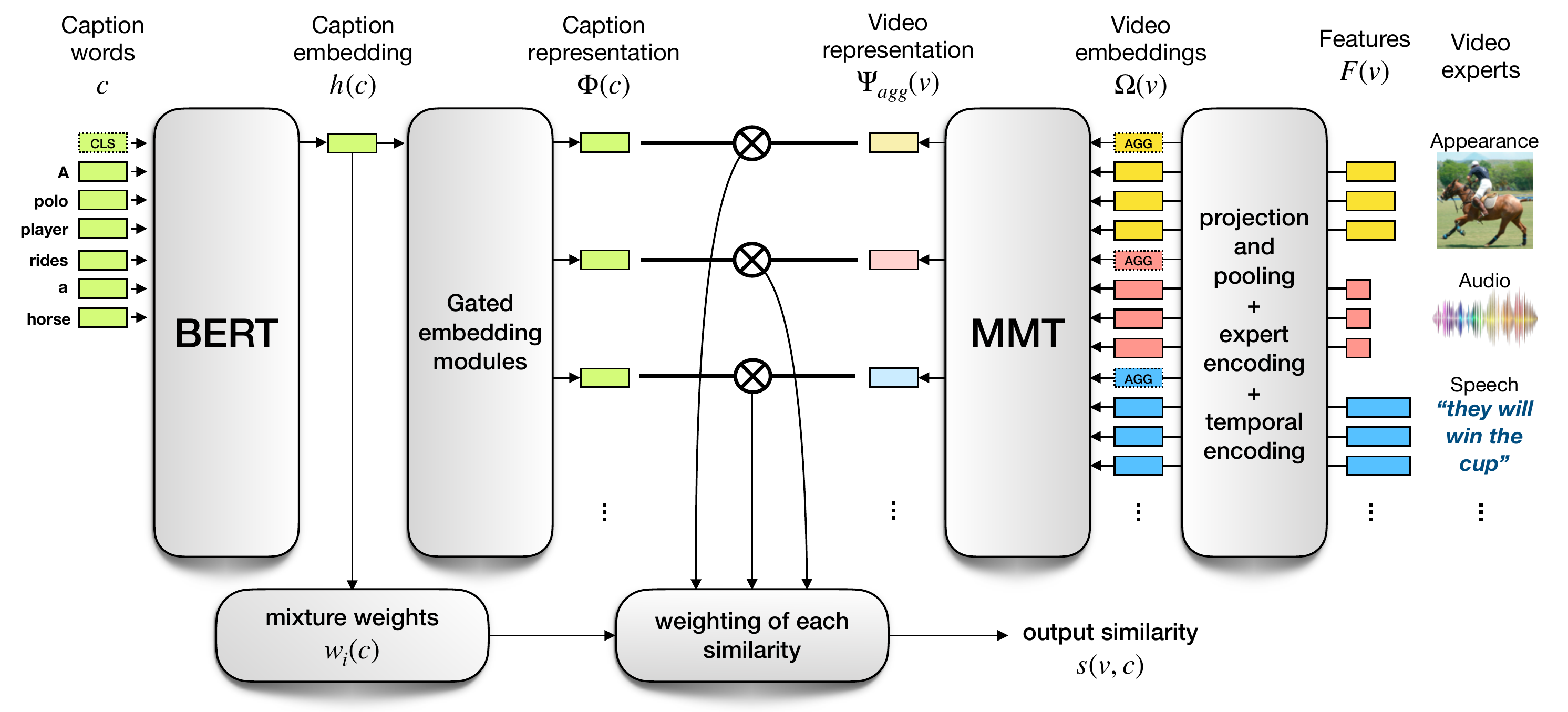}
\end{center}
   \caption{The overall framework of the winner's proposed approach. They used Multi-modal Transformer (MMT, right) to encode video, and BERT (left) for text.}
   \label{fig:first_model}
\end{figure*}

\begin{figure*}[ht]
\begin{center}
   \includegraphics[width=0.8\linewidth]{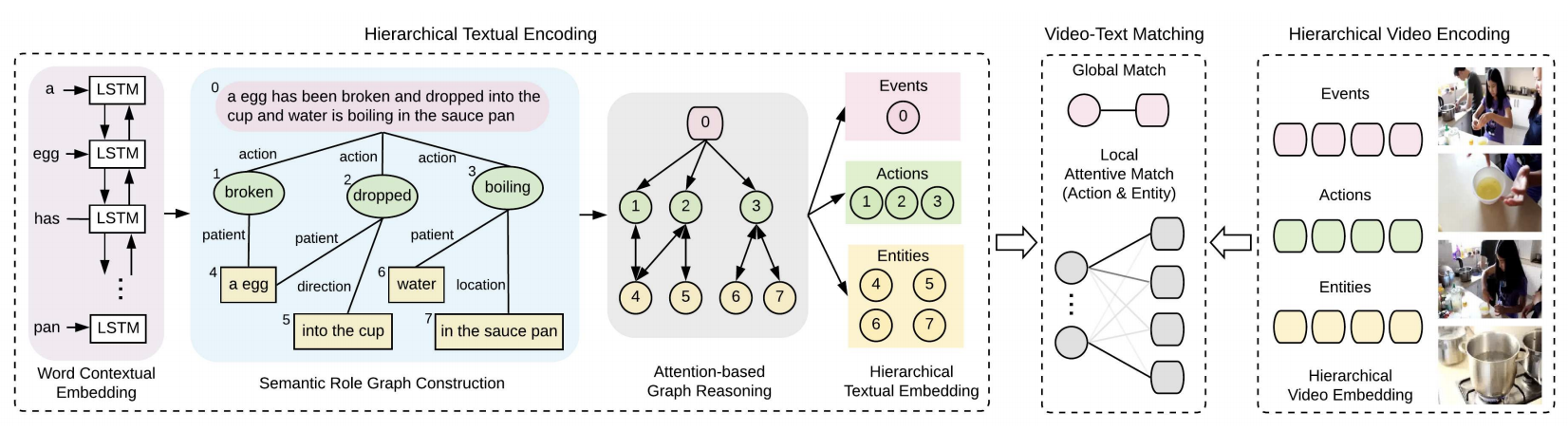}
\end{center}
   \caption{The overall framework of the second place proposed approach -- hierarchical graph reasoning model.}
   \label{fig:second_model}
\end{figure*}

\begin{figure*}[ht]
\begin{center}
   \includegraphics[width=0.8\linewidth]{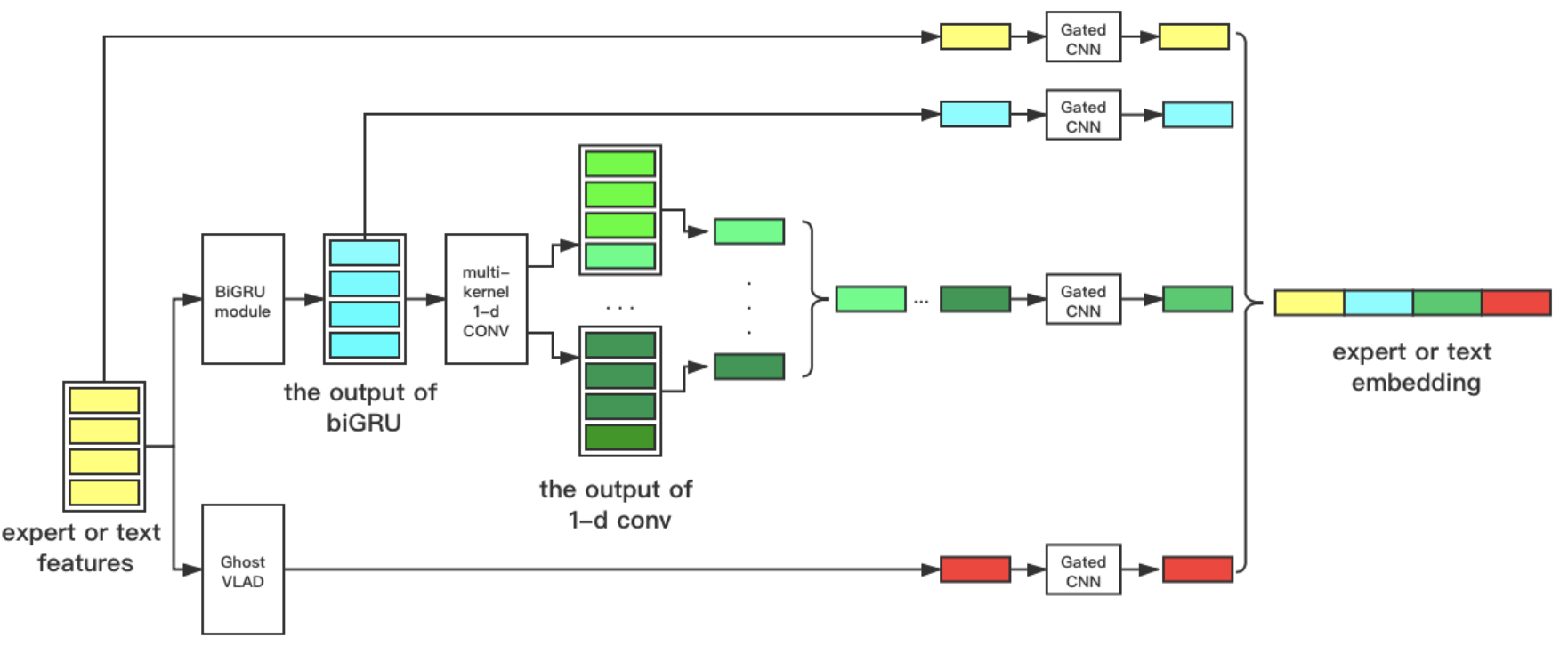}
\end{center}
   \caption{The overall framework of the third place proposed approach -- a hybrid sequence encoder.}
   \label{fig:third_model}
\end{figure*}

\section{Challenge methods and teams} \label{sec:method}

The video understanding pentathlon challenge received 56 submissions from 10 teams in total. The evolution of the leaderboard on the val partition is shown in Fig.~\ref{fig:evolution}. 
Table~\ref{table:final_result} reports the scores using all metrics on the final test partition for each team. Of these, the 4 top teams have declared their affiliation and submitted technical reports. In this document, we provide a brief introduction to the technical reports in order of their overall rank on the public leaderboard. Please refer to the technical reports~\footnote{The technical reports are available at \url{https://www.robots.ox.ac.uk/~vgg/challenges/video-pentathlon/}} for more details. 

\begin{table*}[h]
\hspace{-0.3cm}
\footnotesize
 \begin{tabular}{l|l|c|c|c|c|c|c|c|c}
 \hline \hline 
 Team & Members & Loss & LM & Ensemble \# & Cross-dataset & Temporal agg. & Expert agg. & QE & HM\\
 \hline
 1. MMT           & Valentin Gabeur & Max-Margin & Pretrained & 16 & Yes & Transformer & Transformer & Yes & No \\
Inria             & Chen Sun & Ranking Loss & BERT  &  &  & +Max pool & +MEE & & \\
Google AI         & Karteek Alahari &  &  &  &  &  &  &  & \\
                  & Cordelia Schmid & &  &  &  &  &  &  & \\
\hline
 2. cszhe                & Shizhe Chen & Inverted Softmax  & Glove & 5 & Yes & HGR & HGR & Yes & Yes\\
    Renmin               &  Yida Zhao & +Max-Margin  & +BiLSTM &  & (MSR-VTT) &  &  &   & \\
    Uni. of China. &  Qin Jin & Ranking loss & +HGR  &  &  &  &  &   & \\
\hline 
3. LEgGOdt            & Kaixu Cui & Max-Margin & OpenAI GPT & 1 & Yes & N.A. & Concat & Yes & No \\
   Xinhua  Zhiyun    & Hui Liu & Ranking loss & + BiGRU &  &  &  &  &  & \\
   Tech. Co. Ltd.& Chen Wang &  & +GhostVLAD &  &  &  &  &   & \\
                       & Yudong Jiang &  & +1D-Conv  &  &  &  &  &    &\\
 \hline \hline 
 \end{tabular}
\caption{A summary of the methods from the Top-3 winning teams in the Video Understanding Pentathlon challenge 2020 with the participants' names and affiliations. LM: Language Model, agg.: Aggregation, QE: Query Expansion. HM: Hubness mitigation. Ensemble \#: Ensemble Size}
\label{table:winner_name} 
\end{table*}

Table \ref{table:winner_name} details the winners of the video understanding pentathlon challenge 2020, announced as part of The End-of-End-to-End: A Video Understanding Pentathlon workshop at CVPR 2020.
% A summary of the key features used on each method can also be found in Table~\ref{table:winner_name}. 

\textbf{Rank 1: MMT} is the top-ranking entry by INRIA and Google. The overall framework of their proposed approach is shown in Fig. \ref{fig:first_model}. The team used a multi-modal transformer to jointly encode different video modalities which allowed each of them to attend to the others. The features were then augmented with an expert type encoding and a temporal position encoding. To encode text, they investigated how to jointly optimize the language embedding together with the multi-modal transformer. Team MMT ensembled 16 models for each dataset for their final submission. 
A more detailed study of the method is given in the conference paper version of the method \cite{gabeur:hal-02903209}.

\textbf{Rank 2: cszhe} is the second ranking entry by Renmin University of China. Firstly, the team proposed a hierarchical graph reasoning model~\cite{chen2020fine} which decomposed video-text matching into hierarchical levels for fine-grained retrieval.  The overall framework of the proposed  hierarchical graph reasoning model is shown in Fig. \ref{fig:second_model}. Secondly, they explored query expansion and hubness mitigation methods (by using an \textit{Inverted Softmax}~\cite{smith2017offline}) during the inference to improve a naive nearest neighbor search. Thirdly, they demonstrated that it is beneficial to use additional datasets in a simple multi-task training approach. For the final submission, 3 - 5 models were ensembled for each dataset. 

\textbf{Rank 3: LEgGOdt} is the third ranking entry by Xinhua Zhiyun Technology Co. Ltd.  The team proposed a hybrid sequence encoder in combination with collaborative experts \cite{Liu19a} to construct a common space for the video retrieval task via multi-modal common space learning. The overall framework of the hybrid sequence encoder is shown in Fig.~\ref{fig:third_model}. During training, they trained jointly on all datasets and selected the best performance model for each dataset, and then fine-tuned on each datasets for the final submission.

\textbf{Rank 4: haoxiaoshuai} is the fourth ranking entry by Chinese Academy of Sciences. The team designed a new bi-directional hard-negative ranking loss (Bi-HNRL) that emphasizes on the hardest negatives in the training stage. Specially, they focused on the hardest negative video and query sentence (closest to a positive pair) instead of summing over all negatives.

% \section{2020 Challenge Winners} \label{sec:results}\

\section{Conclusion} \label{sec:conclusion}

We introduced a new Video Understanding Pentathlon challenge at CVPR 2020. The results of the challenge were announced at a Video Understanding Workshop at CVPR, which was also accompanied by invited keynote and spotlight talks. 

\section{Affiliations}

\noindent \textit{Visual Geometry Group, Univ. of Oxford}: Samuel Albanie, Yang Liu, Arsha Nagrani, Ernesto Coto, Andrew Zisserman.  \textit{Inria}: Antoine Miech, Ivan Laptev, Karteek Alahari, Valentin Gabeur, Cordelia Schmid.  \textit{Google}: Rahul Sukthankar, Valentin Gabeur, Cordelia Schmid, Chen Sun. \textit{IVUL, KAUST}: Bernard Ghanem, \textit{Renmin Univ. of China}: Shizhe Chen, Yida Zhao, Qin Jin, \textit{Xinhua Zhiyun Tech Co. Ltd.} Kaixu Cui, Hui Liu, Chen Wang, Yudong Jiang. \textit{Chinese Academy of Sciences}. Xiaoshuai Hao.
\section{Acknowledgements}

The organisers would like to express their gratitude to the creators of the original datasets used in this challenge. They would like to thank in particular Juan Carlos Niebles, Ranjay Krishna, Luowei Zhou, Lisa Ann Hendricks, Jun Xu, Tao Mei, Ting Yao, Yong Rui, David L. Chen, Bryan Russell and Anna Rohrbach for their assistance. We gratefully acknowledge the support of the Programme Grant Seebibyte
EP/M013774/1.

% \samuel{AZ - is there a particular grant that would be appropriate to add? ** Yes, add Seebibyte -- funding for Ernesto, you, etc **}
%%%%%%%%% BODY TEXT

{\small
\bibliographystyle{ieee_fullname}
\bibliography{egbib}
}

\end{document}